\documentclass[sigconf]{acmart}
\usepackage{graphicx}
\usepackage{amsmath}
\usepackage{enumitem}
\usepackage{multirow}
\usepackage{bm}
\usepackage{subfigure}
\usepackage{bbding}
\usepackage{booktabs}
\usepackage{epstopdf}
\usepackage{enumitem}
\usepackage{algorithm}  
\usepackage{algpseudocode}   
\usepackage{balance}
\usepackage{amssymb}
\usepackage{svg}
\usepackage[english]{babel}

\AtBeginDocument{%
	\providecommand\BibTeX{{%
			\normalfont B\kern-0.5em{\scshape i\kern-0.25em b}\kern-0.8em\TeX}}}

\copyrightyear{2025}
\acmYear{2025}
\setcopyright{acmlicensed}\acmConference[KDD '25]{Proceedings of the 31st ACM
SIGKDD Conference on Knowledge Discovery and Data Mining V.2}{August 3--7,
2025}{Toronto, ON, Canada}
\acmBooktitle{Proceedings of the 31st ACM SIGKDD Conference on Knowledge
Discovery and Data Mining V.2 (KDD '25), August 3--7, 2025, Toronto, ON, Canada}
\acmDOI{10.1145/3711896.3736929}
\acmISBN{979-8-4007-1454-2/2025/08}

\begin{document}

\title{Empowering Economic Simulation for Massively Multiplayer Online Games through Generative Agent-Based Modeling}
\author{Bihan Xu$^\dag$}
\thanks{$^\dag$ co-first authors}
\affiliation{%
\institution{State Key Laboratory of Cognitive Intelligence, University of Science and Technology of China}
\city{Hefei}
\country{China}
}
\email{xbh0720@mail.ustc.edu.cn}
\orcid{0009-0005-0163-3322}

\author{Shiwei Zhao$^\dag$}
\affiliation{
\institution{NetEase Fuxi AI Lab}
\city{Hangzhou}
\country{China}
}
\email{zhaoshiwei@corp.netease.com}
\orcid{0000-0002-1017-5897}

\author{Runze Wu*}
\affiliation{
\institution{NetEase Fuxi AI Lab}
\city{Hangzhou}
\country{China}
}
\email{wurunze1@corp.netease.com}
\orcid{0000-0002-6986-5825}

\author{Zhenya Huang*}
\thanks{* corresponding authors}
\affiliation{%
\institution{State Key Laboratory of Cognitive Intelligence, University of Science and Technology of China
}
\country{}
}
\affiliation{%
\institution{Institute of Artificial Intelligence, Hefei Comprehensive National Science Centerce
}
\city{Hefei}
\country{China}
}
\email{huangzhy@ustc.edu.cn}
\orcid{0000-0003-1661-0420}

\author{Jiawei Wang}
\affiliation{%
\institution{Hangzhou Institute for Advanced Study, University of Chinese Academy of Sciences}
\city{Hangzhou}
\country{China}
}
\email{wangjiawei231@mails.ucas.ac.cn}
\orcid{0009-0009-0091-3866}

\author{Zhipeng Hu}
\affiliation{
\institution{NetEase Fuxi AI Lab}
\city{Hangzhou}
\country{China}
}
\email{zphu@corp.netease.com}
\orcid{0000-0003-4367-0816}

\author{Kai Wang}
\affiliation{
\institution{NetEase Fuxi AI Lab}
\city{Hangzhou}
\country{China}
}
\email{wangkai02@corp.netease.com}
\orcid{0000-0002-7767-2329}

\author{Haoyu Liu}
\affiliation{
\institution{NetEase Fuxi AI Lab}
\city{Hangzhou}
\country{China}
}
\email{liuhaoyu03@corp.netease.com}
\orcid{0000-0002-8998-1217}

\author{Tangjie Lv}
\affiliation{
\institution{NetEase Fuxi AI Lab}
\city{Hangzhou}
\country{China}
}
\email{hzlvtangjie@corp.netease.com}
\orcid{0000-0001-9858-809X}

\author{Le Li}
\affiliation{
\institution{NetEase Fuxi AI Lab}
\city{Hangzhou}
\country{China}
}
\email{lile@corp.netease.com}
\orcid{0009-0001-2749-9150}

\author{Changjie Fan}
\affiliation{
\institution{NetEase Fuxi AI Lab}
\city{Hangzhou}
\country{China}
}
\email{fanchangjie@corp.netease.com}
\orcid{0000-0001-5420-0516}

\author{Xin Tong}
\affiliation{%
 \institution{National University of Singapore}
\country{Singapore}
 }
 \email{xin.t.tong@nus.edu.sg}
 \orcid{0000-0002-8124-612X}

\author{Jiangze Han}
\affiliation{%
 \institution{University of British Columbia}
\city{Vancouver}
\country{Canada}
 }
 \email{han.jiangze@gmail.com}
 \orcid{0000-0001-7396-8475}

\renewcommand{\shortauthors}{Bihan Xu et al.}

\renewcommand{\shorttitle}{LLM-based Agents for Simulating Game Economics}



\begin{abstract}
Within the domain of Massively  Multiplayer Online (MMO) economy research, Agent-Based Modeling (ABM) has emerged as a robust tool for analyzing game economics, evolving from rule-based agents to decision-making agents enhanced by reinforcement learning. Nevertheless, existing works encounter significant challenges when attempting to emulate human-like economic activities among agents, particularly regarding agent reliability, sociability, and interpretability.  
In this study, we take a preliminary step in introducing a novel approach using Large Language Models (LLMs) in MMO economy simulation.  Leveraging LLMs’ role-playing proficiency, generative capacity, and reasoning aptitude, we design LLM-driven agents with human-like decision-making and adaptability. These agents are equipped with the abilities of role-playing, perception, memory, and reasoning, addressing the aforementioned challenges effectively. Simulation experiments focusing on in-game economic activities demonstrate that LLM-empowered agents can promote emergent phenomena like role specialization and price fluctuations in line with market rules.
\end{abstract}

\ccsdesc[500]{Computing methodologies~Simulation environments}
\ccsdesc[300]{Information systems~Decision support systems}


\keywords{Massively Multiplayer Online Game, Game Economics, Large Language Model, Generative Agent-Based Modeling}


\maketitle
\newcommand\kddavailabilityurl{https://doi.org/10.5281/zenodo.15526912}

\ifdefempty{\kddavailabilityurl}{}{
\begingroup\small\noindent\raggedright\textbf{KDD Availability Link:}\\
The source code of this paper has been made publicly available at \url{\kddavailabilityurl}.
\endgroup
}
\section{Introduction} \label{intro}
\begin{figure}
    \centering
     \includegraphics[width=0.99\linewidth]{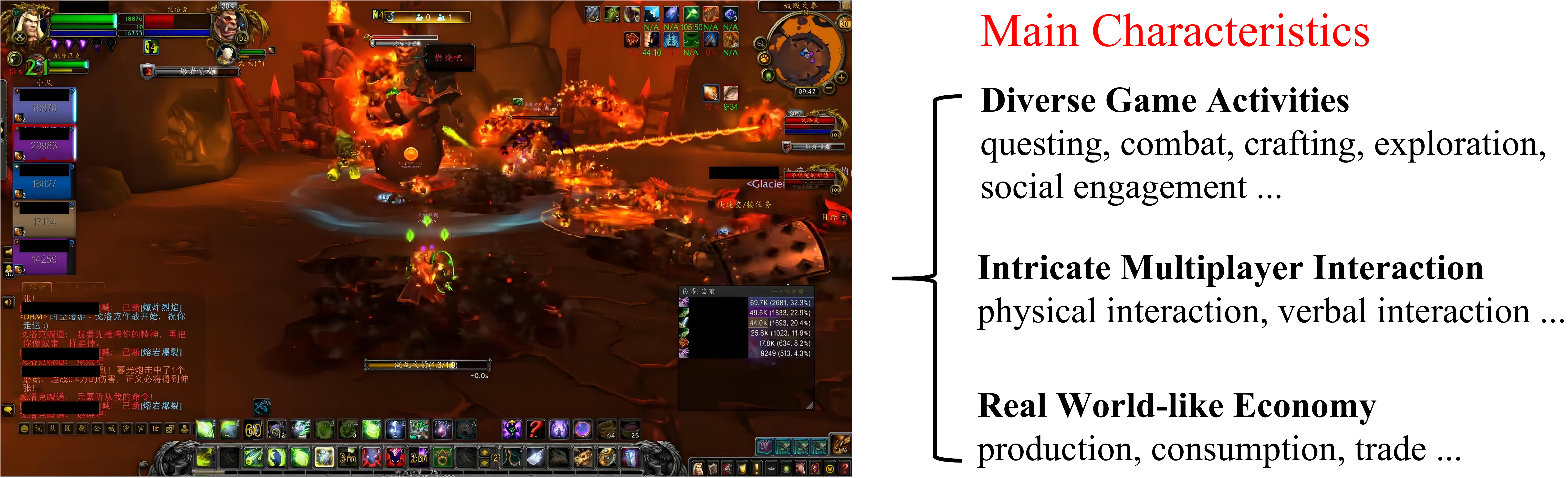}
     \vspace{-0.3cm}
    \caption{Example of the popular MMO game, World of Warcraft, and its key characteristics.}  
    \vspace{-0.6cm}
    \label{fig:intro1}
\end{figure}

The market for Massively Multiplayer Online games (MMOs) reach\-ed \$11.4 billion in 2023 and is projected to hit \$20.36 billion by 2030, at a Compound Annual Growth Rate (CAGR) of 8.2\% between 2024 and 2030\footnote{\url{https://reports.valuates.com/market-reports/QYRE-Auto-16A2286/global-mmo-games}}. 
MMOs are distinguished within the online video game landscape by their capacity to facilitate large-scale player interactions within a persistent and dynamically evolving virtual world~\cite{milik2017persona,hu2023deep}. 
Exemplary MMO examples like World of Warcraft\footnote{\url{https://worldofwarcraft.blizzard.com/}} 
provide players with comprehensive in-game progression activities spanning questing, combat, crafting, and social engagement, as depicted in Fig.~\ref{fig:intro1}.
Central to the MMO player experience is the seamless integration of multifaceted gameplay with the intricate virtual economies, meticulously crafted to emulate the real world and enhance immersion~\cite{kaminski2006impacts,wilkinson2011economic}.
Players within MMOs participate in economic chains that include production, consumption, and trade, deeply experiencing economic realities such as supply and demand dynamics, inflation, and even downturns.
However, formulating robust economic policies in these ecosystems is often constrained by the infeasibility of counterfactual data and experimental testing~\cite{schreiber2021game}. This underscores the need for advanced analytical tools, with economic simulations~\cite{zhao2024mmo,stephens2021measuring,rupp2024geevo} playing a pivotal role. These tools enable policymakers to model and evaluate the impacts of policy interventions within complex economic frameworks.

\begin{figure}[t]
  \centering
  \includegraphics[width=1.0\linewidth]{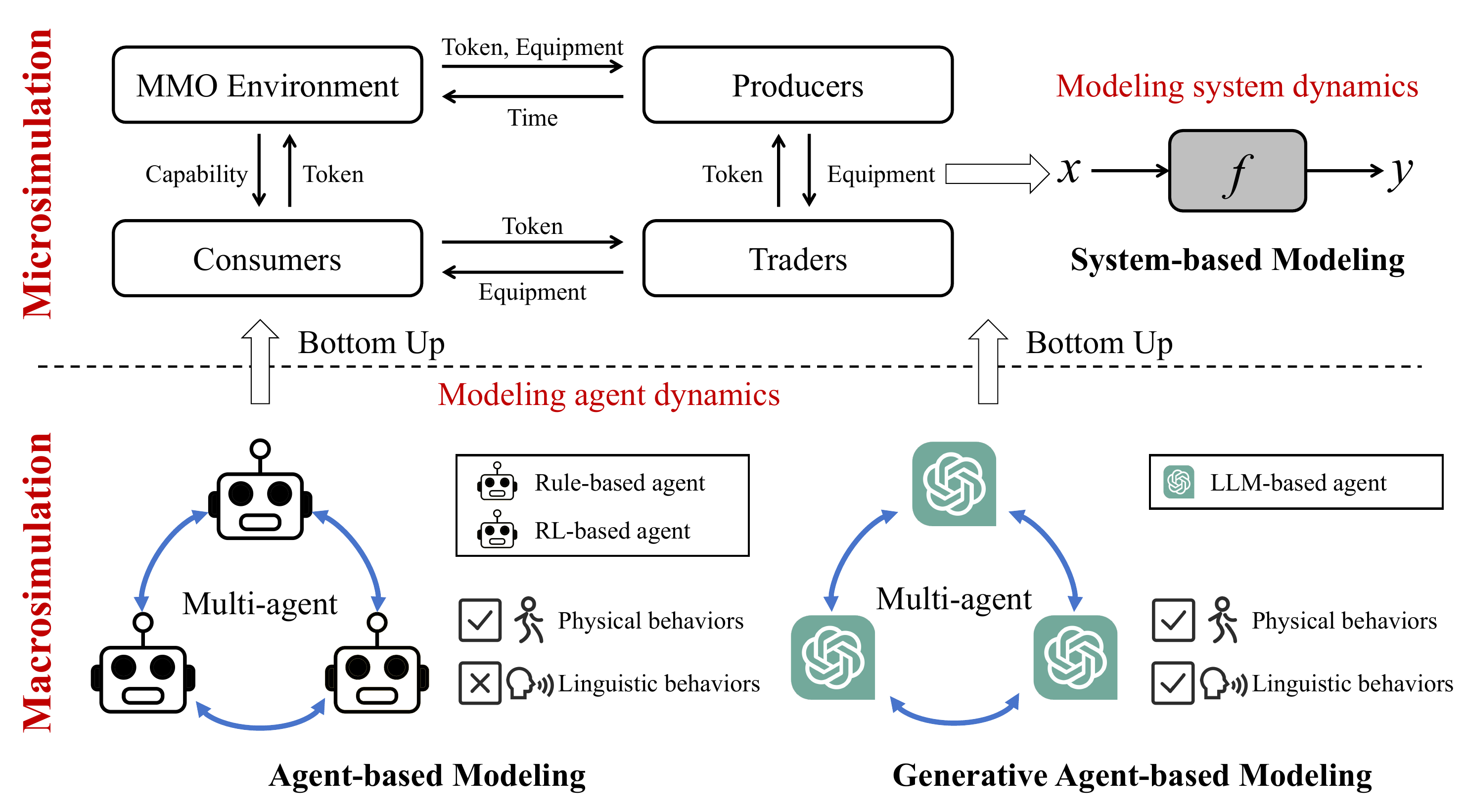}
 \vspace{-0.6cm}
  \caption{SBM and ABM simulation in MMO economies. SBM focuses on specific economic chain links, using complex functions to model the macroeconomic dynamics at the systemic level. Conversely, ABM adopts a bottom-up approach, examining the macro phenomena emerging from microeconomic interactions among individual agents to provide granular insights into macroeconomic pattern formation.}
  \label{sys}
   \vspace{-0.4cm}
\end{figure}

In the realm of economic simulation, two methodological paradig\-ms predominate: System-Based Modeling (SBM) and Agent-Based Modeling (ABM). As shown in Fig.~\ref{sys}, these paradigms offer complementary insights into the dynamics of complex systems, focusing on the macro and micro perspectives, respectively~\cite{ahmadi2015system,batkovskiy2015study,axtell2022agent}.
Given MMO economies' intricate and open nature, ABM has become particularly favored. It emphasizes fundamental agents and encompasses techniques ranging from rule-based \cite{zook2019monte,imagawa2015enhancements,devlin2016combining,lin2006integrating,lin2006enhancing} to Reinforcement Learning (RL)-based approaches \cite{gudmundsson2018human,zhao2024mmo,stephens2021measuring}.
RL-based agents, like the MMO Economist~\cite{zhao2024mmo}, have notably advanced the fidelity of replicating individual economic behaviors within gaming ecosystems. However, several limitations persist:
\begin{itemize}[leftmargin=*,itemsep=1.0pt]
\item \textbf{Reliability.} The RL paradigm uses tailored reward functions to mimic diverse human behavior. However, capturing the demographic diversity of game players in mathematical models is a formidable task.  Moreover, as ABM is a bottom-up method that observes macro-level outcomes from micro-level interactions, depicted in Fig.~\ref{sys}, any deviation from the micro-level realities (i.e., player-level behaviors) within the game can undermine the simulation's credibility.
\item \textbf{Sociability.} A critical yet often neglected aspect of RL is the ability of agents to communicate and interact directly, especially through linguistic mechanisms like bargaining and negotiation. These interpersonal dynamics are fundamental to in-game economic transactions and significantly influence the MMO economic landscape. Neglecting these interactions can result in simulations that overlook the nuanced social dynamics essential to these virtual economies.
\item \textbf{Interpretability.} 
The unclear decision-making mechanisms inherent to RL-based agents hide why they take actions, making it harder to understand how the simulated economy operates. This hinders the use of simulation tools in practical decision-making processes, contradicting their intended utility for policymakers.
\end{itemize}

In recent years, the advancement of Large Language Models (LLMs) has given rise to generative agents that convincingly mimic human behavior for interactive applications~\cite{wang2024survey,xi2023rise,ruan2023tptu,gao2025agent4edu}. One promising approach is to augment traditional ABM agents with LLM integration, known as Generative ABM~(GABM)~\cite{park2023generative,li2023large,arsanjani2013spatiotemporal}. 
This novel approach holds potential for managing the intricacies of MMO economic simulations with specialized capabilities.

\begin{itemize}[leftmargin=*,itemsep=1.0pt]
\item \textbf{Role-playing Proficiency.} 
LLMs have shown strong role-playi\-ng capabilities~\cite{shen2024hugginggpt,zeng2023evaluating,li2023large}, enabling them to make decisions based on character described in texts. This simplifies agent construction in ABM and allows for creating agent profiles directly from real player data, enhancing authenticity in simulations.
\item \textbf{Generative Capacity.} 
LLMs' advanced text skills enable agents to use natural language to understand and express intentions \cite{jin2023data,liang2023encouraging,xi2023rise,li2023camel,qian2023communicative,zhao2024comprehensive}, even implicit ones, in complex economic interactions. This fills a crucial gap in MMO economic simulations, capturing psychological and sociological aspects akin to real-world human interactions.
\item \textbf{Reasoning Aptitude.} 
LLMs' reasoning abilities help them explain decision-making rationales \cite{xue2024decompose,madaan2024self,liu2025know,wei2022chain,zhou2022least,li2025foundation,ma2025debate}. This enhances the interpretability of agent behaviors, providing developers with deeper insights into the motivations and strategic considerations underlying agents' actions.
\end{itemize}
Leveraging LLMs' capabilities, we propose an LLM-empowered simulation framework for MMO economies to tackle the aforementioned limitations:
(1) We employ a data-driven approach to construct agent profiles from real-world data, thereby bridging the gap between real-world scenarios and simulation environments.
(2) The public and private chat scenarios are integrated to facilitate communication and information flow among agents, thereby embedding socialization into the economic simulation.
(3) We design the LLM-empowered MMOAgent, comprising five components: profile, perception, reasoning, memory, and action. This agent is endowed with the capacity to comprehend, explore, and interact within the game environment, akin to human players.
Our main contributions are summarized as follows:
\begin{itemize}[leftmargin=*]
    \item We have pioneered the integration of LLMs with ABM in MMO economic simulation, addressing the limitations of traditional ABM approaches. This integration introduces generative ABM, a new paradigm for studying and analyzing MMO economies.
    \item Following the open-source work of the MMO Economist~\cite{zhao2024mmo}, we enhance it with full player-to-player (P2P) trading,  featuring direct linguistic negotiation and bargaining between players which is absent in previous work. This extension bridges a critical gap in MMO economics, enabling more realistic economic interactions.
    \item Building upon this foundation, we have developed a sophisticated generative agent, termed MMOAgent. This agent is designed to comprehensively understand and navigate the MMO economic system, mirroring real human behavior patterns. 
    \item  
    The experimental results consistently replicate real-world patterns, such as the agents' role specialization in economic chains and game resource pricing fluctuations that align with market principles. This underscores its excellence in both advanced research and practical applications within the gaming industry.
\end{itemize}

\begin{figure*}[t]
  \centering
  \includegraphics[width=0.8\linewidth]{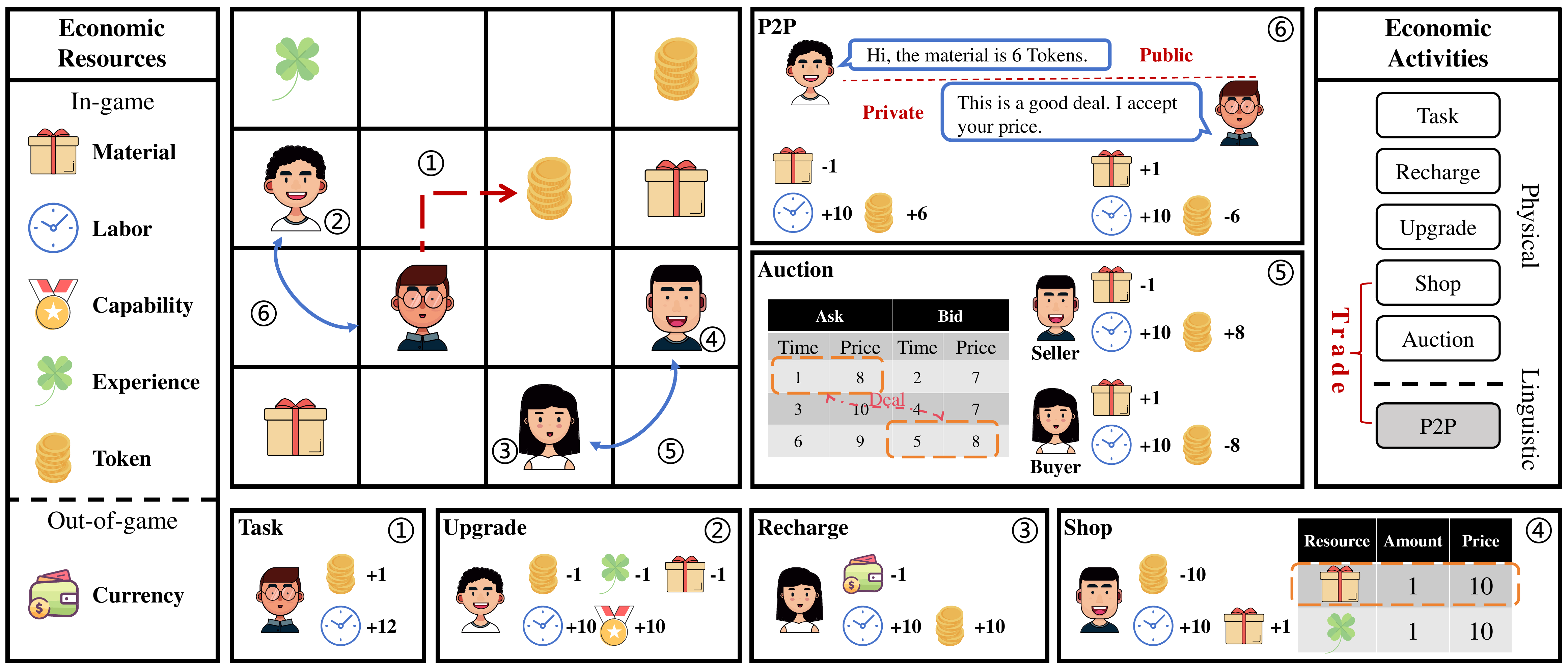}
  \vspace{-0.3cm}
  \caption{An overview of our expanded virtual environment for MMO economies.}
  \label{fig:environ}
   \vspace{-0.5cm}
\end{figure*}

\section{Related Work}
\subsection{Simulation in MMO Economy}
In the realm of economic simulation, two prevailing methodological paradigms are SBM and ABM, serving as pivotal tools for analyzing, predicting, and optimizing intricate economic phenomena and policies~\cite{batkovskiy2015study,axtell2022agent,eisen2017simulating}. SBM employs mathematical models to simulate the entire economic system, focusing solely on macro-level phenomena, albeit requiring assumptions that often contradict real-world conditions~\cite{batkovskiy2015study}. In contrast, ABM focuses on individual micro-level information and serves as a promising alternative for simulating systems from the bottom up~\cite{axtell2022agent}. Given the complexity and open nature of MMOs, economic models within these games often utilize ABM for implementation~\cite{devlin2016combining,bjornsson2009cadiaplayer,li2010multi,lin2006integrating,lin2006enhancing}. 
Li et al.~\cite{li2010multi} introduced a rule-based multi-agent simulation model to analyze dominant entities' behavior in a supply chain and assess factors influencing decision uncertainty. 
Stephens and Exton~\cite{stephens2021measuring} propose an RL framework to evaluate virtual economies by simulating MMO inflation and analyzing in-game resource price dynamics and economic vulnerabilities. The MMO Economist~\cite{zhao2024mmo} creates a holistic simulation environment for personalized agents' diverse economic activities, serving as a foundation to formulate adaptive adjustment strategies for resource allocation between grinding and in-game purchases to balance profitability and equity.

In conclusion, MMO economic simulations offer a robust framework for dissecting complex economic dynamics within MMOs.
Nevertheless, these methods have their limitations. Rule-based simulations are often too simplistic to capture the intricate interactions among players, while RL-based simulations are less interpretable. Moreover, they are not as effective in customizing players' profiles and modeling verbal social interactions. Consequently, this paper incorporates LLMs to address the aforementioned issues.

\subsection{LLM Empowered Multi-Agent Simulation}
The autonomous agent, empowered by LLMs, has garnered substantial attention in recent years \cite{wang2024survey,xi2023rise,ruan2023tptu,liu2024socraticlm}.  
An innovative application lies in the construction of a simulation environment populated by agents for emergent social phenomena learning ~\cite{park2023generative,li2023large,arsanjani2013spatiotemporal,jinxin2023cgmi,argyle2023out,wang2024user,li2023you}. 
For example, Generative Agents~\cite{park2023generative} and AgentSims~\cite{lin2023agentsims} constructed an AI town with LLM-empowered agents acting as genuine citizens, engaging in daily tasks, news sharing, and social networking. This offers insights into information dissemination among agents. $S^3$ \cite{gao2023s} utilized real-world social network data to condition LLM-empowered agents, emulating individual and collective behaviors in social settings to study the spread of emotions on issues like gender discrimination.
Li et al. \cite{li2023large} simulated the work and consumption behaviors of LLM-empowered agents in a macroeconomic context, demonstrating more rational and stable macroeconomic indicators compared to traditional approaches.
Additionally, LLM agent-based simulation has been successfully applied in some gaming environments, where LLM exhibits great reasoning and instruction following capabilities, enabling seamless gameplay in games such as Civilization and Werewolf \cite{qi2024civrealm,xu2023exploring,wu2024enhance,wang2023voyager}. However, there is still a dearth of research in the context of more intricate MMO gaming environments, notably within the domain of MMO economic systems, which remains largely unexplored.

\section{Virtual Environment for MMO Economies}\label{sec:env}
As shown in Fig.~\ref{fig:environ}, the MMO Economist~\cite{zhao2024mmo} has designed a sophisticated virtual environment that emulates the intricate economic structures of MMOs. It covers various in-game activities, including resource allocation, market transactions, fiscal policies, and in-game purchases, offering a robust platform for the empirical assessment of diverse economic policies and their ramifications on the in-game economy. The environment integrates a variety of economic resources such as experience (EXP), material (MAT), tokens (TOK), currency (CCY), capability (CAP), and labor (LAB), and a range of economic activities like tasks, upgrades, auctions, shops, and recharges. More details can be found in \textit{Appendix~\ref{desp_eco}}.


\textbf{Limitations.}
The aforementioned virtual environment, while innovative, exhibits two fundamental limitations. First, its transactional mechanisms are limited to pseudo-P2P~(player-to-player) trading, mediated by either the game environment encompassing Non-Player Characters~(NPCs) or intermediary systems like auctions. This setup fails to capture the essence of direct player interactions in genuine transactions~\cite{green2022science,ali2023sequential}. Second, the emphasized economic activities primarily focus on physical behaviors, revolving around the "production-consumption-trade-forex" economic chains, thereby overlooking the pivotal impact of linguistic behaviors within the economy. Direct P2P trading, prevalent in MMOs, involves two key features~\cite{bilir2009real}: (1) players engage in transactions without any game mediation; (2) these transactions closely parallel real-world economic practices where verbal negotiation and bargaining play a central role. Consequently, the current virtual environment cannot simulate P2P trading in MMO economies. 

\begin{figure*}[t]
  \centering
  \includegraphics[width=0.83\linewidth]{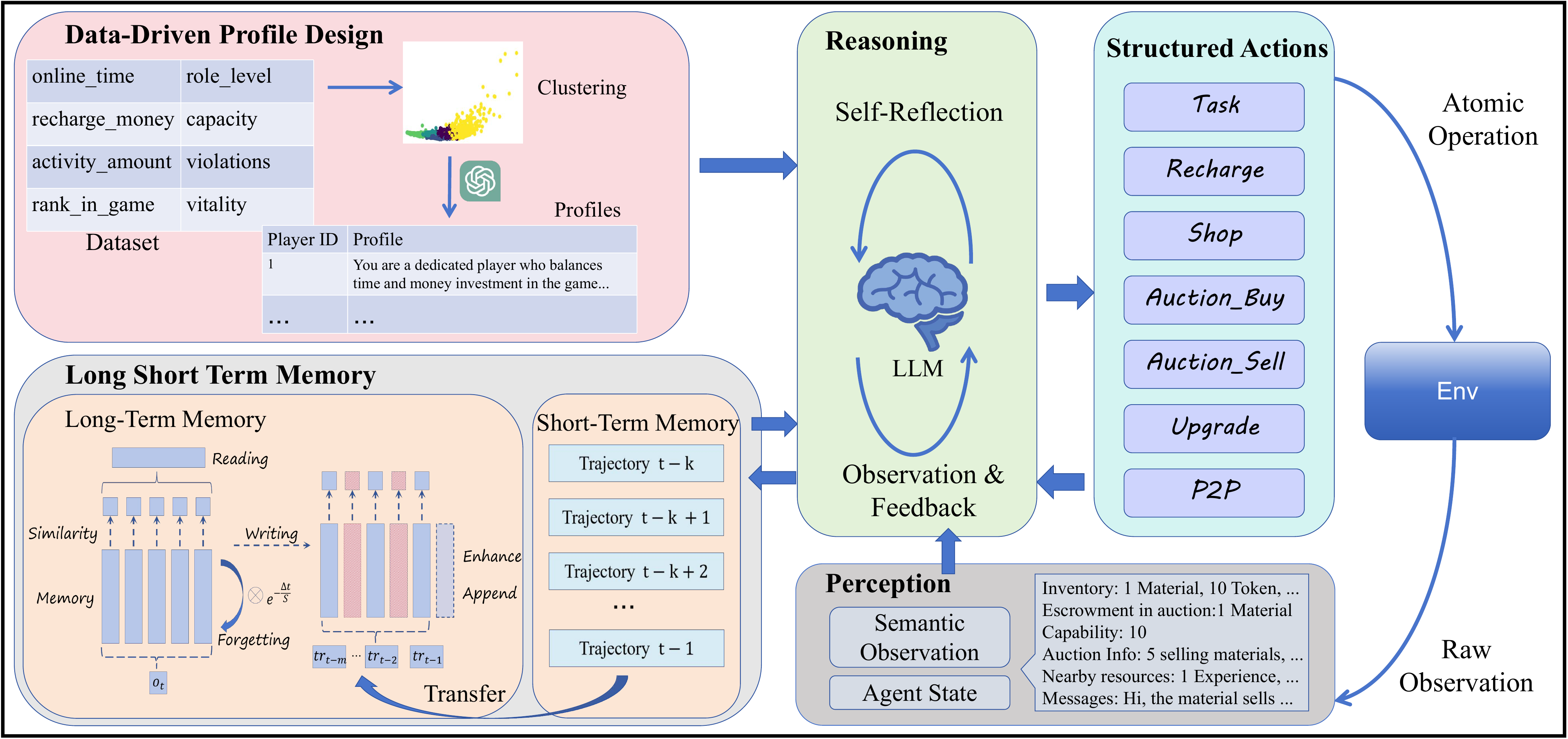}
  \vspace{-0.2cm}
  \caption{An overview of our GABM simulation framework for MMO economies.}
  \label{fig:model}
   \vspace{-0.4cm}
\end{figure*}

\textbf{Extensions.}
To address this gap, our research expands the virtual environment to facilitate P2P trading, incorporating two distinct online communication modalities depicted in Fig.~\ref{fig:environ}:  (1) Public Chatting, enables players to send messages visible to all players, creating a communal marketplace for broadcasting trade offers. (2) Private Chatting,  allows players to send messages visible to a specific player, facilitating private negotiations between interested buyers and sellers.
These two chatting mechanisms allow agents to use verbal behavior such as negotiation and bargaining~\cite{fu2023improving,abdelnabi2023llm} to accomplish genuine P2P trading. Both buyers and sellers will utilize selling and bidding orders in the auction for price negotiations.
These extensions introduce a new paradigm for direct private transactions, thereby filling the research gap left by previous studies.

\section{Framework}
As shown in Fig. \ref{fig:model}, we present the MMOAgent, an LLM-empowered framework engineered to simulate micro-level activities for MMO economies. It consolidates five core modules: profile, perception, reasoning, memory, and action.  The profile module tailors the agent to mirror real player traits. The perception module interprets the game environment, while the reasoning module determines suitable action at each timestep. The memory module logs game experience from previous steps, and the action module executes permissible actions within the game. By integrating these modules, MMOAgent is capable of comprehending, exploring, and interacting with the game environment, akin to human players.

\subsection{Data-driven Profile Design}
\label{profile_design}
In game economics, creating realistic agents is paramount for the bottom-up simulation methodologies (e.g. ABM), which investigate macro phenomena emerging from micro-level interactions among individuals. 
Previous studies~\cite{li2010multi,zhao2024mmo}, by standardizing agents or controlling them with artificial rewards, have overlooked the intrinsic complexity required for agents to authentically emulate human behavior. This oversight limits their anthropomorphism and makes the simulations less dependable.

To bridge real-world contexts and simulated environments, we use authentic player data to create personalized profiles for simulated players.  Initially, we collect detailed player records from an anonymous MMO released by NetEase Games\footnote{\url{https://game.163.com/}}, under the premise of ensuring data privacy. These player records encompass player demographics, payment details, historical behavior, and more, as outlined in \textit{Appendix~\ref{features}}. Next, using the k-means clustering method \cite{hartigan1979algorithm}, we group various player traits and identify the centroids of all clusters to yield k representative player characteristics.
Finally, GPT-4~\cite{achiam2023gpt} is leveraged to generate personalized player profiles with text descriptions for each set of player characteristics as detailed in \textit{Appendix \ref{profile_text}}. To accommodate GPT-4's limited sensitivity to numerical data~\cite{yu2023temporal}, each characteristic is divided into five distinct levels: high, medium-high, medium, medium-low, and low. This process allows for a more detailed understanding of player behavior in simulated environments.
The generated profiles will be allocated to the simulating agents based on probabilities determined by the ratio of the cluster size relative to the total population.

\subsection{Perception}
To help LLM-powered agents understand the game environment better, we integrate the perception module with a parser. This parser processes complex raw observations from the environment into text for comprehension. 
Specifically, the parsed observation of the agent in Fig.~\ref{fig:model} encompasses the following textual representations:
\vspace{-0.1cm}
\begin{itemize}[leftmargin=*]
    \item \textbf{Inventory} pertains to the numerical count of economic resources held by the player.
    \item \textbf{Escrowment in auction} refers to the player's economic assets placed in the auction for sale.
    \item \textbf{Auction info} includes details of the selling and bidding orders with their prices in the auction.
    \item \textbf{Nearby resources} shows the spatial distribution of nearby resources in the environment.
    \item \textbf{Messages} denotes communications received from other players via public or private chatting.
\end{itemize}
\vspace{-0.1cm}
\subsection{Structured Actions with Execution Feedback}\label{structured action}
Traditional RL methods derive decisions directly from the game's atomic action space, such as simple movements (e.g., moving up on the game map), which can result in a lack of logical connection between consecutive
decisions. This contrasts with LLMs, which possess advanced reasoning and planning abilities but lack precise control over low-level task-specific operations \cite{zhu2023ghost}.
To simplify decision-making for LLMs among numerous low-level operations, we encapsulate the economic activities into well-defined functions with clear semantics (see \textit{Appendix \ref{action_desp}}).  These functions reconcile the LLM's cognitive abilities with the agent's low-level control needs. For example, in the "Task" of resource collection, RL methods select movement directions at each step. Conversely, the LLM employs predefined functions to translate the cognitive "Task" decision into movement sequences, using DFS algorithm for resource exploration and A* for shortest path generation, as depicted in Fig. \ref{fig:environ}. 

However, some selected actions may fail if the agent's inventory doesn't meet requirements. 
Therefore, we construct a verifier based on game rules to provide feedback on executed actions, signaling success or failure. In case of failure, it explains. For instance, if the agent lacks currency, attempting a recharge will yield a failure message, explicitly stating the currency insufficiency as the reason. This failure message will serve as valuable feedback for the LLM reasoning, enabling it to recognize its error and subsequently adjust the decision-making process accordingly.

\subsection{Feedback Enhanced Reasoning and Planning}
\subsubsection{Feedback Enhanced Reasoning.} 
By delineating the structured actions with precise semantics and functionalities, an LLM-based reasoning module is employed to determine appropriate structured actions via environmental observations to accomplish predetermined objectives. This ensures both cognitive decision-making and successful action implementation in the agent's environment.
Specifically, we use the zero-shot chain-of-thought prompting technique \cite{wei2022chain} to encourage the LLMs to carry out more reasoning before deciding. As Section \ref{structured action} notes, LLMs may repeat failed plans without feedback due to outcome unawareness. Therefore, to avoid generating unchecked actions, the reasoning module will reevaluate the action using the execution feedback as follows:
\begin{equation}
 \centering
    \small
    a_t = CoT(pf, obs_t, mem_t, fb, \Theta).\label{cot}
\end{equation}
where $pf$ signifies the agent's profile. $obs_t$ is the current observation. $mem_t$ means the past trajectories and experience in memory, to be elaborated in Section \ref{memory}. $\Theta$ is the parameters of LLM. 


\subsubsection{Reflection and Future Planning.}
Developing effective game strategies is crucial for success. For example, in resource-rich environments, optimal strategies prioritize maximizing resource collection via diverse tasks, while in scarce scenarios, focusing on trading or recharging becomes preferable. Agents must therefore perceive long-term environmental changes and adapt strategies accordingly, mirroring human adaptability and enabling continuous learning. To achieve this, we implement a periodic reflection mechanism: every $n$ step, the agent evaluates the $n$ most recent actions and environmental observations to devise a new game strategy for subsequent gameplay. The reflection and strategy are generated as follows:
\begin{equation}
 \centering
    \small
sr = CoT(as, os, sr_{\text{prev}}, \Theta), \
\end{equation}
where $as$ and $os$ represent the sequence of $n$ previous executed actions and corresponding observations, respectively. $sr_{\text{prev}}$ is the previous self-reflection text. Current reflection $sr$ will be stored in the memory and utilized for decision-making in the next period.

\subsection{Numeric-aware Long Short Term Memory}\label{memory}

In game environments, decision-making relies heavily on past experiences and environmental shifts. Integrating a memory module is essential for agents to grasp game dynamics and make informed decisions. We implement two specialized memory modules: the Short-Term Memory (STM) captures the intricate details of an agent's immediate past, while the Long-Term Memory (LTM) emphasizes the experiences that contribute to success on a higher level.
 
\subsubsection{Short-Term Memory.}
 To dynamically capture the recent movements and behaviors of an agent, we employ the STM module that stores the agent's trajectory data. Specifically, 
 at each step t, a record $<obs_t, a_t>$ is inserted into the STM, capturing the environmental observation and the corresponding action taken by the agent. To ensure temporal relevance, the STM is structured to retain only the ten most recent trajectories, all of which are then utilized to aid subsequent decision-making within the context of LLMs.
 
\subsubsection{Numeric-aware Long-Term Memory.}
In addition to STM, which just records recent trajectories, LTM is proposed to store valuable game experiences over extended periods. A record stored in LTM is represented as $<obs_i, a_i, s_i>$ and $s_i$ denotes its importance score.

\textbf{Memory Embedding.}
Games offer a rigorously numerical setting with dynamic fluctuations in various values (e.g. agent's inventory), pivotal for numerically informed decision-making. Traditional semantic representations of past observations \cite{park2023generative,wang2024user} 
 often miss these numerical dynamics in gaming contexts. To address this, we consolidate the agent's observations into a unified representation through the concatenation of various numerical values.

\begin{table*}[t]
    \setlength{\abovecaptionskip}{0pt}
	\centering
	\caption{Results of baselines and ablation experiments on simulating player's performance in the game. Existing state-of-the-art results are underlined and the best results are bold. We compare our MMOAgent with the SOTA MMO-economist and $\bf{^\ast}$ indicates p-value $< 0.05$ in the t-test.}
	\renewcommand\arraystretch{1.0}
    \normalsize
	{
		\begin{tabular}{l|c|c|c|c|c|c}
            \hline
			\multirow{2}[3]{*}{Methods}   & \multicolumn{2}{c|}{Rich}  & \multicolumn{2}{c|}{Moderate}  & \multicolumn{2}{c}{Scarce}\\
   \cmidrule{2-7} 
			          & Capability & Diversity   & Capability & Diversity   & Capability & Diversity    \\
    \hline	
			Random & 35.0 & 2.5631 & 34.0 & 2.5649 & 31.0 & 2.5663     \\
			Rule-based & 84.0 & 2.0541 & 68.0 & 2.0625 & \underline{58.4} & 1.9774 	\\
               MMO-economist & \underline{92.4} & 1.8893 & \underline{72.6} & 1.7767 & {51.8} & 1.6485   \\
			ReAct  & 47.0 & 0.7624 & 37.4 & 0.7309 & 32.0 & 0.8139  \\
			Reflexion & 51.0 & 0.7860 & 39.2 & 0.7583 & 35.2 & 0.8092			 \\
			
			MMOAgent (GPT-3.5) & \textbf{121.0$\bf{^\ast}$} & 1.5822 & \textbf{80.4$\bf{^\ast}$} & 1.4149 & \textbf{75.0$\bf{^\ast}$} & 1.3106     \\
			
            MMOAgent (Llama3) & 104.0 & 1.1582 & 76.2 & 1.0156 & 68.4 & 0.9850\\ 
   \hline
			
	
			MMO w/o STM & 115.2 & 1.4845 & 77.0 & 1.4045 & 72.2 & 1.2961 \\	
			MMO w/o LTM & 108.4 & 1.4459 & 74.6 & 1.3525 & 68.0 & 1.2525  \\	
			MMO w/o Reflect & 112.6 & 1.4496 & 76.8 & 1.3742 & 65.4 & 1.1582   \\	
			
            \hline
		\end{tabular}%
	}
	\vspace{-0.3cm}
	\label{t3}%
\end{table*}

\textbf{Memory Reading.}
When the agent makes a decision, it initially extracts pertinent information from its memories, which are relevant to the current observation. 
The similarity between two records, x and y, is determined by the numerical variance of their respective numerical embeddings:
\begin{equation}
    \centering
    \small
    sim(x, y)  = 1 - \max_{i} \frac{|x_i - y_i|}{\max(x_i, y_i)},
\end{equation}
where $x_i$ and $y_i$ denote the $i$-th elements in embedding vectors. The agent then identifies the most suitable memory reference for the current situation by balancing similarity and importance.
\begin{equation}
 \centering
    \small
    reference = \mathop{\arg\max}_{i \in LTM} sim(emb_t,emb_i) + s_{i}.
\end{equation}
In decision-making, the retrieved experience, along with STM and self-reflection text, is referenced as $mem_t$ in Eq. \eqref{cot}.

\textbf{Memory Writing.}
Individuals are greatly impacted by previous successes, serving as vital experiences for subsequent decision-making. Consequently, we emphasize documenting significant past actions that contribute to the player's overall performance. Following the MMO Economist \cite{zhao2024mmo}, we utilize the utility function to evaluate the reward of each action. For a positive reward action (e.g., Upgrade), we focus on its preceding $m$ closest trajectories stored in STM and employ a recency-sensitive scoring mechanism to measure their contributions.
\begin{equation}
 \centering
    \small
    s_i = r_t * \gamma^{t - i},\quad\quad i = t - m, \dots, t - 1, \label{eq:write}
\end{equation}
where $r_t$ denotes the reward of action $a_t$, 
and $\gamma$ is the discount factor.
Then, these $m$ meaningful trajectories will be written into the LTM with their importance score as $<obs_i, a_i, s_i>$. 
Particularly, upon adding a trajectory i to the LTM, we compute its similarity with each record in LTM and accumulate its importance scores if the similarity exceeds a predefined threshold.

\textbf{Memory Forgetting.}
Based on cognitive theory~\cite{averell2011form,xu2023learning}, memory fades over time, leading us to reduce the importance score of records to simulate forgetting. Drawing from the Ebbinghaus forgetting curve~\cite{ebbinghaus2013memory,huang2020learning}, we adopt an exponential form to specify this process:
\begin{equation}
 \centering
    \small
    s_i^t = s_i^{t-1} * exp(- \frac{1}{S}), \label{eq:forget}
\end{equation} 
 where S is a hyperparameter that denotes the strength of memory. If a record's importance score is less than a predefined threshold, it will be removed from the memory.

\section{Experiments}
This section commences with a description of the experimental setup followed by quantitative and qualitative evaluations to answer the following research questions (RQs):
RQ1: Can the LLM-based agent fully understand the game environment?
{RQ2}: Is the agent's behavior pattern consistent with the real human?
{RQ3}: How does the simulation system exhibit the economic phenomenon?


\subsection{Simulating Players' Performance Results}
\subsubsection{Experimental Setup}\label{exp}
An agent's ability to fully simulate human behavior relies on a thorough understanding of the game environment and its rules. Without this, credible simulation is impossible. As a result, we first evaluated our MMOAgent's gaming performance against established baselines: 
\begin{itemize}[leftmargin=*]
   \item{\textbf{Random}} is an agent that selects actions from its action space uniformly at random.
    \item{\textbf{Rule-based}} is a rule-based agent implemented by us that utilizes heuristic rules to guide its behavior, such as adaptive action space and resource-oriented navigating algorithms.
    \item{\textbf{MMO-economist}} uses RL to optimize its behavior in MMO economies \cite{zhao2024mmo}.
    \item{\textbf{ReAct}} incorporates chain-of-thought prompting \cite{yao2022react}, generating reasoning traces and action plans using LLMs. Feedback from the environment and agent states are used as observations.
    \item{\textbf{Reflexion}} is built upon the ReAct, which incorporates self-reflection mechanisms to guide future actions \cite{shinn2024reflexion}. We implement a heuristic reflection strategy that prompts the agents to engage in self-reflection if their performance fails to improve over 10 consecutive steps.
\end{itemize}
This comparison spanned three scenario settings, \textbf{Rich}, \textbf{Moderate}, and \textbf{Scarce}, each with a different allocation of game resources, from abundant to scarce. Our assessment hinges on two metrics: \textit{Capability}, which directly correlates with the agent's resource acquisition and management effectiveness, and \textit{Diversity}, reflecting the variety of the agent's chosen activities. See \textit{Appendix \ref{sec:settings}} for more details.

\subsubsection{Performance Comparison}
To assess the MMOAgent's efficacy, we compared its gameplay performance against all baselines, with results detailed in Table \ref{t3}.
We have several key findings. 
Firstly, MMOAgent significantly outperforms all baselines on the \textit{Capability} metric across all scenarios, indicating superior environmental comprehension. 
Secondly, MMOAgent exhibits balanced performance in exploiting and exploring multiple activities as measured by the \textit{Diversity} metric. This sets it apart from Random and Rule-based agents, which, despite showing high diversity, have inferior game capability due to restricted planning capacity. Similarly, MMOAgent differs from ReAct and Reflexion, whose lower diversity correlates with reduced game performance, underscoring their constraints in activity exploration.
Thirdly, MMOAgent surpasses earlier LLM-based agents (i.e., ReAct and Reflexion) which do not fully manage past experiences. It excels through adept use of game insights and advanced planning capabilities.
Lastly, with GPT-3.5 serving as the LLM backbone for MMOAgent, it has consistently outperformed the open-source alternative, Llama3. As a result, subsequent analysis experiments are completed based on GPT-3.5.

\begin{table*}
	\centering
	\normalsize
    \caption{Evaluating scores of the consistency of multiple agents with their assigned profiles. }
    \vspace{-0.3cm}
	\renewcommand\arraystretch{1.1}

	{
		\begin{tabular}{c|c|c|c|c|c}
			\hline
			Agent & 1  & 2  & 3  & 4 & 5   \\ 
   (Type)& (Engaged Grinder) & (Moderate Player) & (Spending Enthusiast) & (Casual Gamer)& (Steady Participant)\\
			\hline
			Consistency &3.89 & 3.79 &3.75  &3.82 &3.86\\
			\hline
		\end{tabular}
	}
	
        \label{human}
        \vspace{-0.3cm}	
\end{table*}

\begin{figure*}
    \centering
    \subfigure[The interpretability of decision making.]{\includegraphics[width=0.42\hsize]{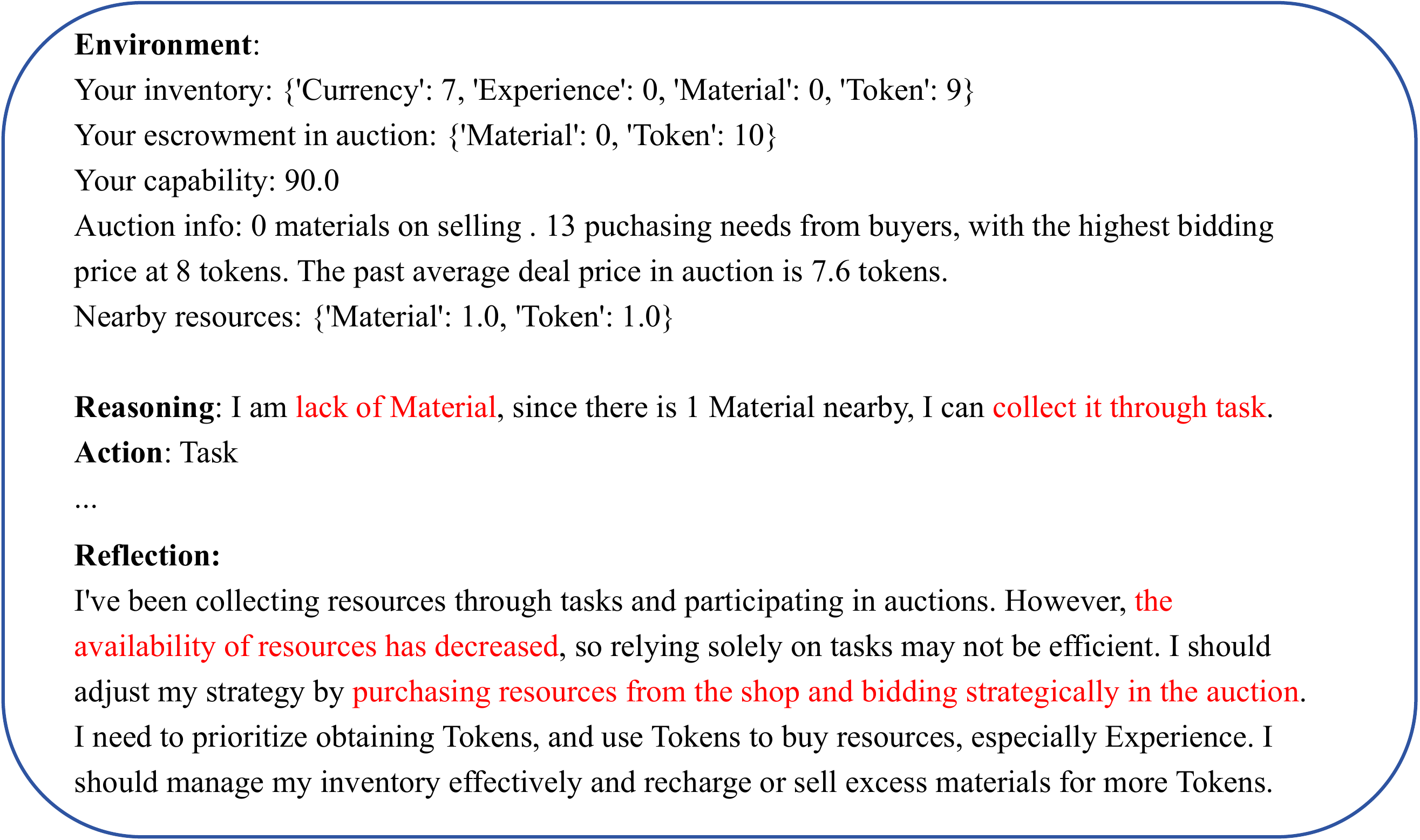} \label{Inter}
    \setlength{\abovecaptionskip}{0pt}}
    \subfigure[The example of P2P trading.]{\includegraphics[width=0.42\hsize]{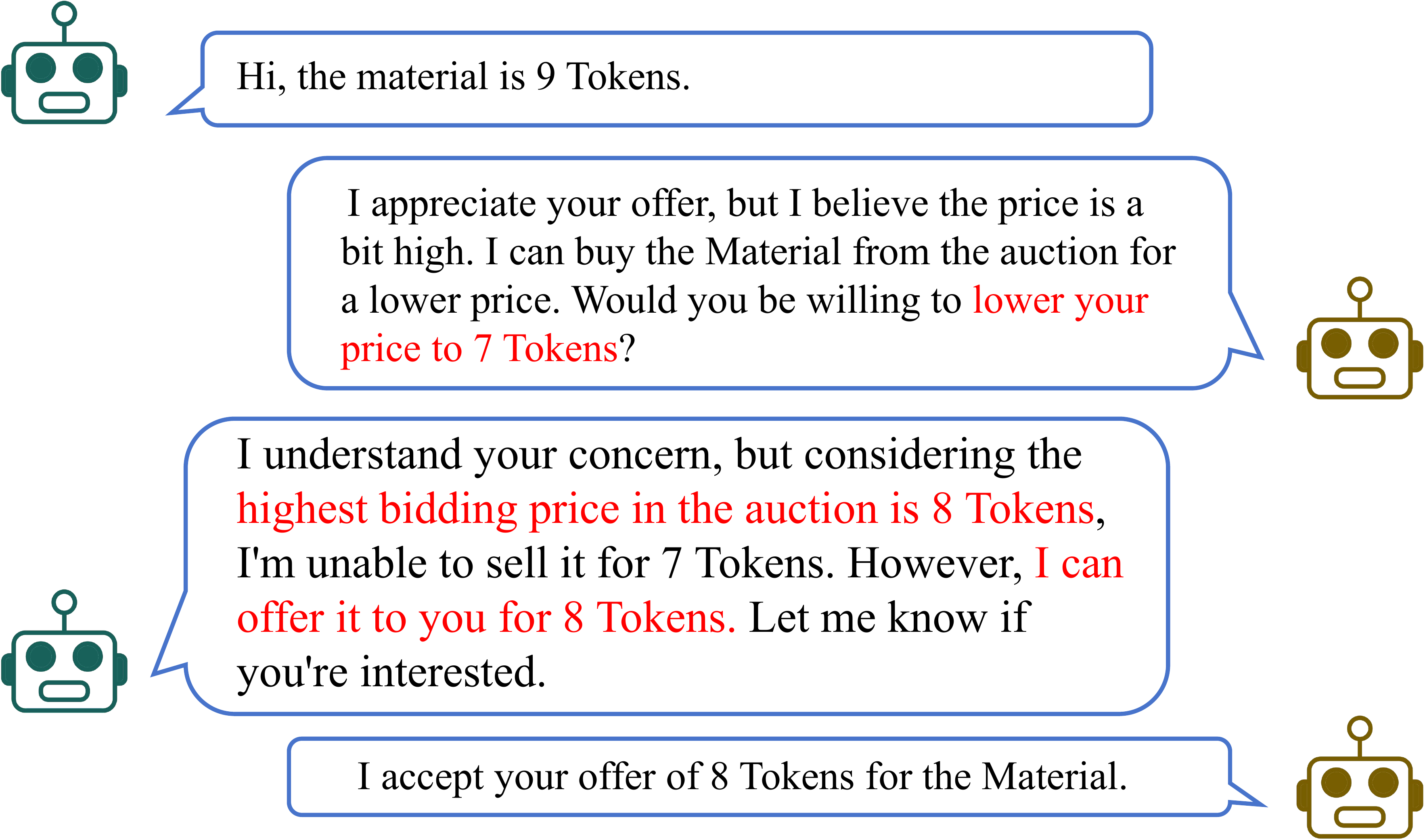} \label{PVP}}
    \vspace{-0.4cm}
    \caption{Case study of decision-making and P2P trading.}
    \vspace{-0.2cm}
\end{figure*}

\subsubsection{Ablation Study} \label{ablation}
We conducted an ablation study on MMOAgent's three key modules, creating following variants: MMO w/o STM, MMO w/o LTM, and MMO w/o Reflect, which respectively remove short-term memory, long-term memory and cycle reflection. Our results in Table \ref{t3} revealed that removing any part of  MMOAgent leads to performance decline, emphasizing the importance of all proposed components. Notably, the absence of LTM degrades performance most in \textbf{Rich} and \textbf{Moderate} scenarios, highlighting the necessity of key experience retention. Additionally, MMO w/o Reflect's decreased performance underscores the value of explicit reflection, while MMO w/o STM's diminished effectiveness emphasizes the need for short-term environmental awareness.

\subsubsection{Interpretability}
Fig. \ref{Inter} illustrates the cognitive processes underlying the decision-making of the agent. Initially, the proximity of available resources justifies the pursuit of resource collection through tasks. As these resources dwindle over time, continued task pursuit becomes less feasible.  Consequently, the agent engages in reflective reasoning and adapts by strategizing resource acquisition through trade. In conclusion, MMOAgent effectively detects environmental changes, dynamically adjusts strategies through cyclical reflection, and provides clear rationale.


\subsection{Human-like Consistency Examination}\label{consis}
\subsubsection{Profile Consistency}
To verify if agents' decisions match their profiles, we created a 5-tier rating system (ranging from 5 for a perfect match to 1 for a total mismatch), as detailed in \textit{Appendix \ref{Evaluate}}. To minimize manual effort and utilize GPT-4's advanced capabilities over GPT-3.5 and its human-like evaluation skills \cite{achiam2023gpt,naismith2023automated,hackl2023gpt,li2023camel}, we primarily used GPT-4 for consistency assessment. To ensure GPT-4's reliability, three human evaluators rated a 20\% sample, achieving 95\% agreement with GPT-4. Thus, GPT-4's results were deemed reliable and used as final outcomes (Table \ref{human}). Notably, the ratings are consistently close to 4, demonstrating that MMOAgent can produce sequential decisions that align well with the inherent characteristics of each designated profile. The score is not close to 5 may be because the degree words like "fair" used in profiles to describe preferences are ambiguous, making it hard for humans or GPT-4 to determine whether a sequence fully matches the corresponding profile. 


\subsubsection{Role Specialization}
Each agent possesses distinctive attribut\-es and inclinations, as delineated in their profiles, and the experiments underscore a clear role specialization. Agents adapt their strategies to align with their individual preferences, as illustrated in Figure \ref{dist}. For instance, cost-averse agents like Agents 1 and 5 tend to engage in labor-intensive activities such as task completion and trading. Notably, Agent 1, who favors trading, actively engages in auctions and P2P, profiting from resource sales to other players, as depicted in Fig. \ref{tranfer}. Conversely, agents willing to spend money evolve into Pay-to-Win players (i.e., Agents 2 and 3), realizing that recharging and purchasing resources proves more profitable. This emergent behavior stems entirely from the agents' diverse characteristics, showing varied economic activity preferences rather than adherence to rigid rules as observed in traditional ABM.
\subsection{Economic Phenomena on System Level}
\subsubsection{Experimental Setup}
To emulate economic phenomena at a systemic level more effectively, we randomly assign 30 agents with profiles based on the representing cluster's size in Section~\ref{profile_design}, and proceed with a 200-step simulation to conduct our analysis.

\subsubsection{Difference in Auction and P2P Trade}
Fig.~\ref{PVP} shows a P2P trade example, highlighting the agent's negotiation skills by integrating auction insights with bargaining strategies. Our analysis of the average transaction prices for MAT reveals 6.86 tokens for auctions and 6.46 tokens for P2P trades. This outcome signifies a notably lower deal price in P2P trades compared to auctions, consistent with their nature of direct, escrow-free personal exchanges.




\subsubsection{Supply and Demand Rule}
\begin{figure*}
    \centering
    \vspace{-0.3cm} 
    \subfigure[Diagram of the price and supply and demand of MAT in the auction.]{
     \includegraphics[width=0.4\linewidth]{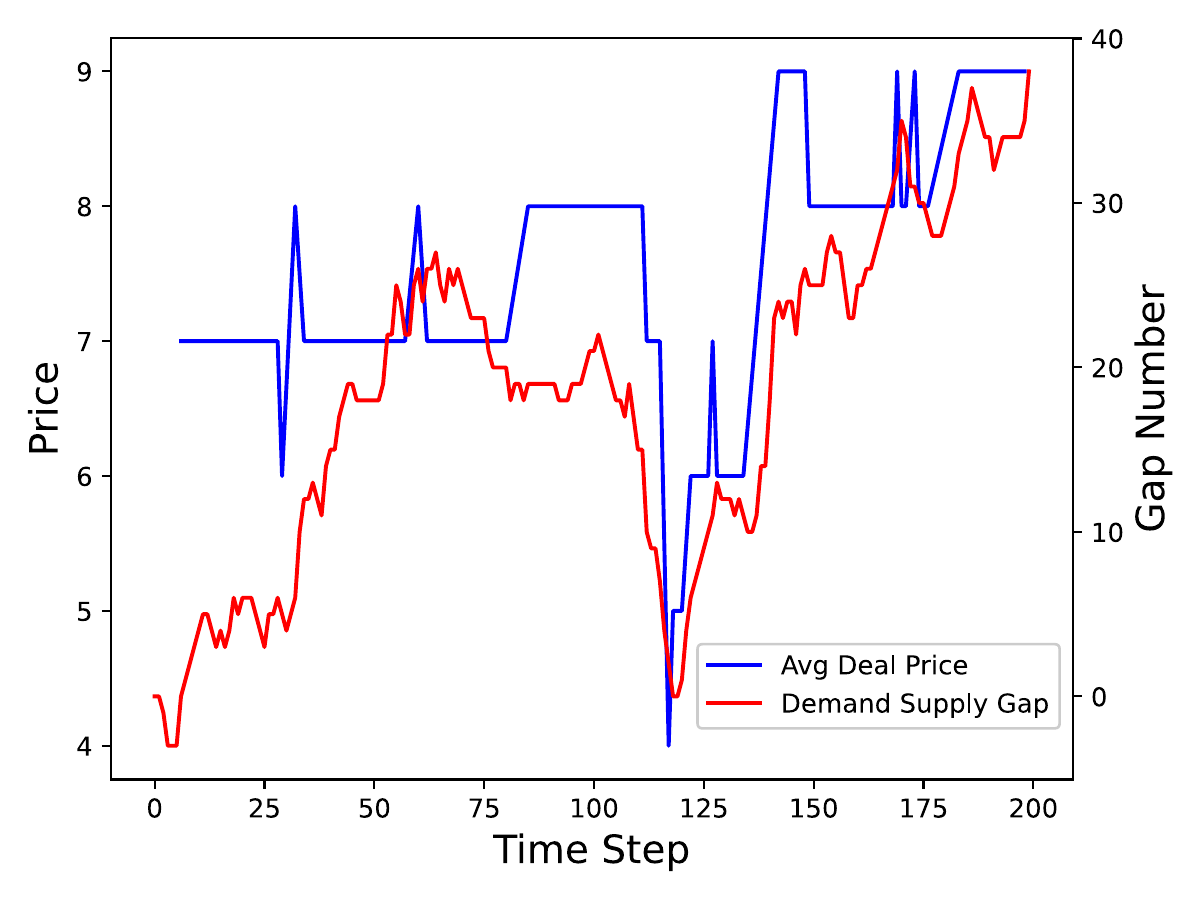}\label{demand_supply}}
    \subfigure[Diagram of the profitability and equality across three scenarios.]{\includegraphics[width=0.4\hsize]{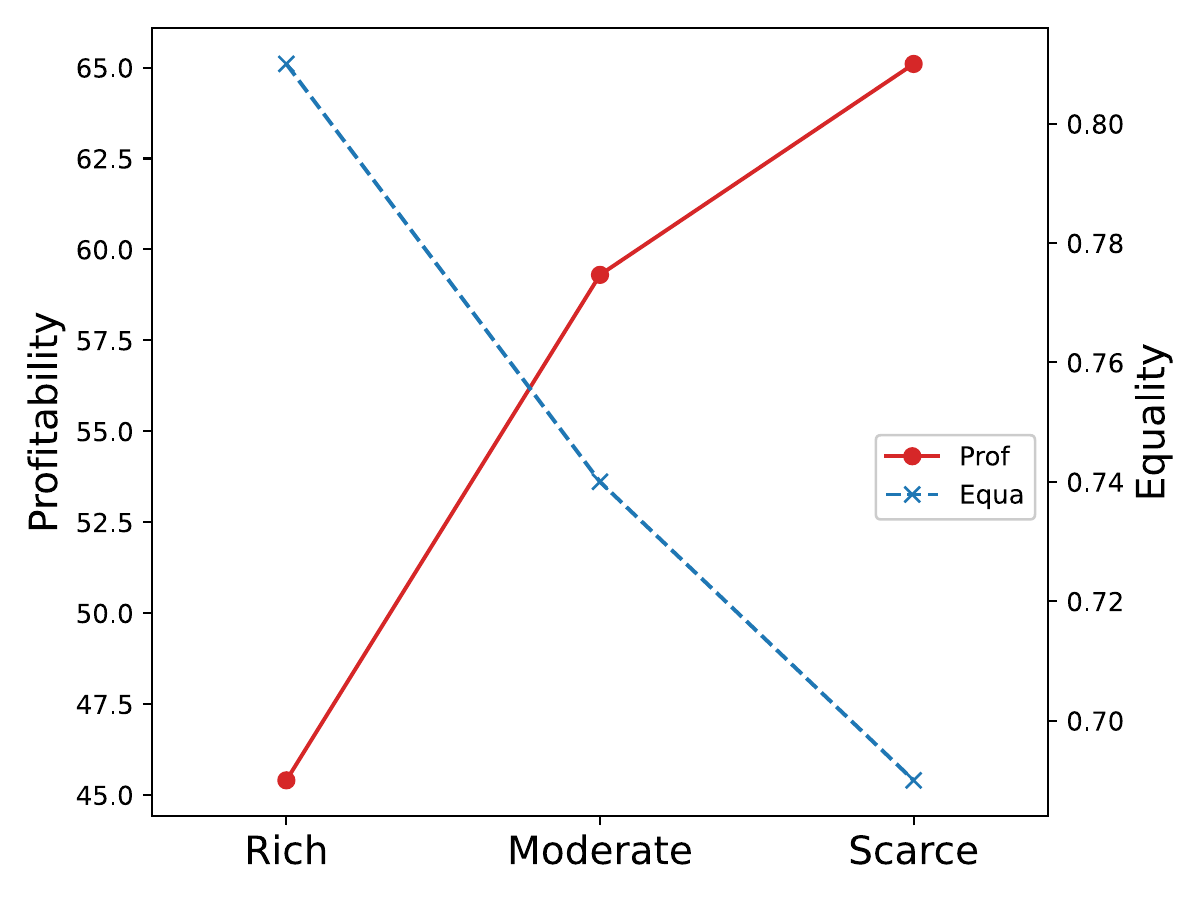} \label{prof_equa}}
    \vspace{-0.5cm}
    \caption{Two system-level emergent economic phenomena.}  
    \vspace{-0.4cm}
\end{figure*}

Auction prices should adhere to fundamental supply and demand principles \cite{gale1955law}, increasing when demand exceeds supply or purchases surpass market rates, and decreasing when supply outweighs demand or offers fall below market rates. Our analysis of auction price fluctuations is detailed in Fig. \ref{demand_supply}.  We introduce the "Demand Supply Gap" variable to represent the disparity between bidding orders (demand) and selling orders (supply) in the auction. This variable indicates an undersupplied market if positive and oversupplied if negative. Utilizing the Pearson correlation coefficient, we assess the relationship between the auction price and the demand-supply disparity, obtaining a coefficient of 0.67 with a significance level of $p < 0.001$. This result underscores a fundamental correlation, affirming that our simulation effectively replicates real-world economic principles.

\begin{figure*}[t]
 \small
  \centering
  \subfigure[The transfer of game resources among agents.]
	{
		\includegraphics[width=0.37\textwidth]{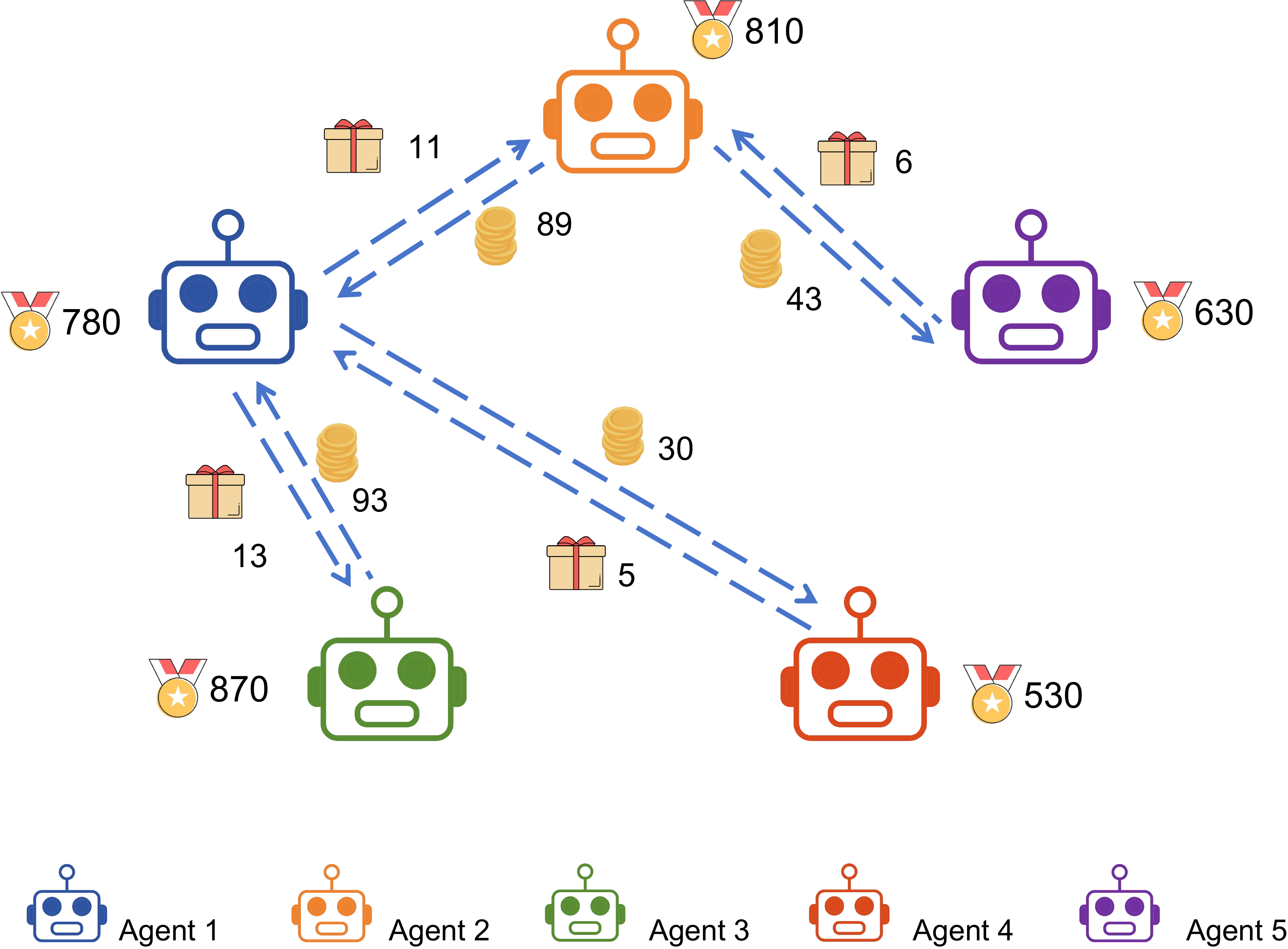}\label{tranfer}
	}
	\subfigure[The action distribution of agents.]
	{
		{\includegraphics[width=0.37\textwidth]{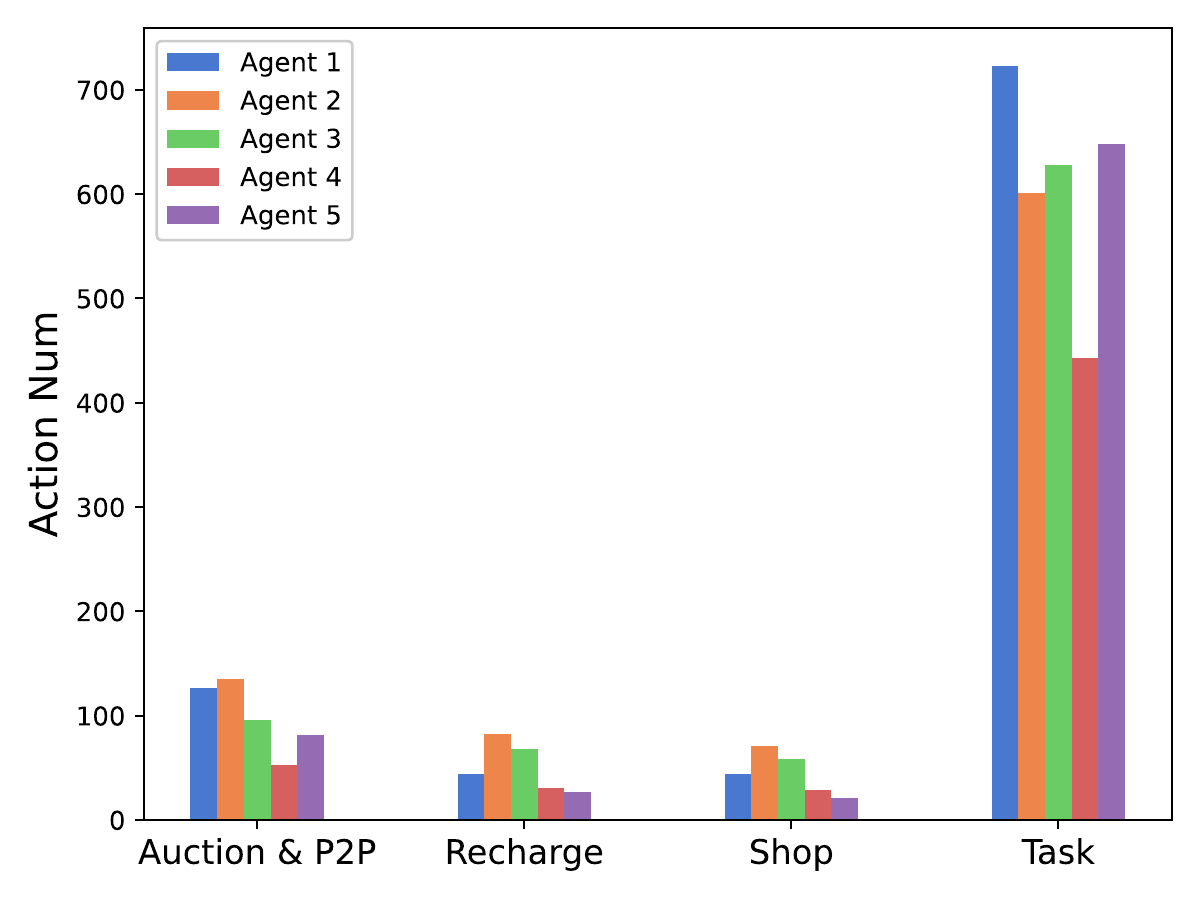}}\label{dist}
	}
 \vspace{-0.5cm}  
  \caption{The role specification of agents.}
   \vspace{-0.3cm}
\end{figure*}

\subsubsection{Equality-Profitability Trade-off Validation}\label{eq_prof}
We employ two system-level metrics, equality, and profitability, to examine the impact of individual decisions on overall system dynamics. These metrics are calculated as proposed in \cite{zhao2024mmo} as follows:
\vspace{-0.2cm}
\begin{equation}
\small
\begin{aligned} 
    & prof = \frac{1}{N}\sum_{i=1}^{N}x_{i,T}^{CCY},
    \quad equa = 1 - Gini({x_{1,T}^{CAP}, \cdots, x_{N,T}^{CAP}})\frac{N}{N - 1},
    \\
    & Gini({x_{1,T}^{CAP}, \cdots, x_{N,T}^{CAP}}) = \frac{\sum_{i=1}^{N}\sum_{j=1}^{N}|x_{i,T}^{CAP} - x_{j,T}^{CAP}|}{2N\sum_{i=1}^{N}x_{i,T}^{CAP}}
\end{aligned}
\end{equation}
where $N$ is the number of players and $x_{i,T}^{CCY}$ is the cumulative amount of currency spent by player $i$ till the end of the simulation $T$. $x_{i,T}^{CAP}$  denotes the cumulative amount of capability created by player $i$. The larger $prof$ means players pay more for better performance. The $equa$, ranging from 0 to 1, measures equality among players, with 1 representing perfect equality and 0 for perfect inequality.
Fig. \ref{prof_equa} presents game profitability and equality across three scenarios. The findings suggest that augmenting resource allocation within the environment promotes equality but diminishes profitability. With ample resources, players can easily gain game assets through tasks, equalizing non-paying and paying players, yet reducing recharge incentives and harming profitability. In contrast, scarce resources will prompt players to recharge more, reversing the dynamics. This demonstrates the MMOAgent's ability to adapt to varying resource distributions, effectively highlighting the trade-off between equality and profitability on the system level.
\section{Limitations and Discussion}
Our research successfully constructs a simulation environment to emulate the dynamics of MMO economies for the purpose of economic strategy formulation and validation. However, due to the inherent hallucination of the LLM, the agent may still generate illegal actions during decision-making, like upgrading without resources or overbidding in auction without enough tokens. Additionally, limited game knowledge leads agents to take conservative actions (i.e., task), reducing action sequence variety (Figure \ref{dist}).
Despite certain limitations, our research holds significant potential. The framework is not confined to MMO trading scenarios. Its modular design—comprising profile, perception, reasoning, memory, action, and reflection mechanisms—mirrors human cognitive processes, making it adaptable to various contexts. By defining the environment, tasks, and action space and adjusting prompts, it can be applied to other contexts like real-world economic simulations. 

\section{Conclusion}
In this paper, we have pioneered the integration of LLMs with ABM in MMO game economic simulations, surpassing the limitations of traditional approaches and presenting a new paradigm for investigating and analyzing MMO economies. We expanded the virtual game environment to encompass full P2P trading rooted in linguistic behavior, addressing a crucial gap in MMO economies. Additionally, we introduced a sophisticated generative agent, MMOAgent, designed to comprehensively understand and navigate the MMO environment, exhibiting behavior patterns resembling those of real humans. Our simulation results within the game environment have demonstrated that this framework accurately mirrors real-world scenarios, including role specialization and fundamental market rules, highlighting its relevance for cutting-edge research and practical applications within the gaming industry.

\begin{acks}
This research was partially supported by the National Natural Science Foundation of China (Grants No.62477044), CCF-NetEase ThunderFire Innovation Research Funding (NO. CCF-Netease 202306), and the Fundamental Research Funds for the Central Universities (No.WK2150110038). Zhenya Huang gratefully acknowledges the support of the Young Elite Scientists Sponsorship Program by CAST (No.2024QNRC001).
\end{acks}

\bibliographystyle{ACM-Reference-Format}
\balance
\bibliography{cite}


\appendix
\section{Economic Resources and Activities}\label{desp_eco}
Six kinds of economic resources form the material foundation and the carrying entity of the economic system as follows:
\begin{itemize}[leftmargin=*]

\item{Experience (EXP)}: Inalienable, intangible assets acquired through gameplay, usable only by individual players and non-tradable.

\item{Material (MAT)}: Tangible assets that are obtained directly from the game and are transferable for player trading.

\item{Token (TOK)}: Universal in-game exchange medium for goods and services, obtainable via gameplay or external recharge.

\item{Currency (CCY)}: Official, government-issued tender from a specific country that is utilized outside of the game environment to facilitate the acquisition of tokens.

\item{Capability (CAP)}: Specific, virtual abilities or scores that are enhanced through active participation in various in-game activities.

\item{Labor (LAB)}: The effort expended by players as they engage in in-game activities during gameplay.
\end{itemize}
Furthermore, five categories of physical economic activities are meticulously designed to encompass the necessary and extendable content of most contemporary MMO economic systems as follows:
\begin{itemize}[leftmargin=*]
\item{Task} broadly refers to any production behavior in which players
directly obtain resources through their labor in the game.
\item{Upgrade} broadly refers to any consuming behavior of players to
improve their capabilities by consuming corresponding economic
resources in the game.
\item{Auction} broadly refers to any trade behavior involving free trade
between players in the game. All tradable economic resources
can be traded structured as a continuous double auction.
\item{Shop} broadly refers to any trade behavior where players directly
purchase commodities from game malls or Non-Player Characters (NPCs) in the game.
\item{Recharge} broadly refers to any forex behavior where players
make payments in the game, such as In-App Purchases~(IAPs).
\end{itemize}

\section{Structured Actions}\label{action_desp}
In Table \ref{structure_actions_t}, we outline structured actions with explicit descriptions. "Rule" indicates applying actions per predefined human regulations. The "Recharge" action costs 1 CCY to obtain 10 TOK per use, while "Shop" enables acquiring the most scarce Upgrade resource from the in-game mall. "Upgrade" requires consuming 1 MAT, 1 EXP, and 1 TOK to increase CAP by 10. For executing complex structured actions, operational capabilities are realized by invoking LLMs, such as determining auction pricing and generating P2P chat responses.

\begin{table}[H]
	\footnotesize
	\renewcommand\arraystretch{1.0}
	\centering
    \caption{Structured actions and descriptions.}
    \vspace{-0.2cm}
	{
		\begin{tabular}{p{1.4cm}|p{4.5cm}|p{1.0cm}}
			\hline
			Actions& Descriptions & Method\\ 
			\hline
			$Task$ & Navigating game map for the nearest resource. & DFS, A* \\
			$Recharge$ & Consuming one currency for tokens. & Rule \\
			$Shop$ & Buy one most lacking resource in game mall.& Rule \\
			$Auction\_Buy$ & Bid on the MAT in the auction. & LLM call \\
			$Auction\_Sell$ & Sell MAT in the auction. & LLM call \\
			$Upgrade$ &Consuming resources to improve Capability. & Rule\\
			$P2P$ & Generate response in the negotiation. & LLM call  \\
			\hline
		\end{tabular}
	}
	\label{structure_actions_t}
         \vspace{-0.6cm}
\end{table}
\begin{table*}
	\footnotesize
	\centering
    \caption{Representative player profile types and corresponding texts.}
    \vspace{-0.2cm}
	{
		\begin{tabular}{p{3cm}|p{12cm}}
			\hline
			Player Type & Profile Text \\ 
			\hline
			Engaged Grinder
 & You are highly active in the game, focusing on gameplay rather than spending money. You may be involved in in-game trading or other activities that don't require financial investment but demand a significant time commitment. (\textcolor{red}{Invest substantial time but little money.})\\
            \hline
			Moderate Player & You are a dedicated player who balances time and money investment in the game. You engage in a fair amount of activity and maintain a strong character. You're willing to spend moderately to enhance your gaming experience but don't rely heavily on recharging to progress. (\textcolor{red}{Invest moderate time and money.}) \\
            \hline
			Spending Enthusiast & You are an enthusiastic player who enjoys a high level of activity and resource management. You're not averse to spending money to advance in the game and maintain a strong character, striking a balance between time investment and in-game purchases. (\textcolor{red}{Invest substantial time and money.}) \\
            \hline
			Casual Gamer & You are a casual player who spends minimal time and money on the game. Your activity and resource levels are modest, and you show little interest in recharging. Gaming is a low-priority leisure activity for you. (\textcolor{red}{Invest minimal time or money.}) \\
            \hline
			Steady Participant & You enjoy playing the game at a steady pace without a significant investment of time or money. Your activity level and resource management are moderate, and you prefer not to spend much on in-game purchases, focusing instead on enjoying the game without rushing. (\textcolor{red}{Invest moderate time but little money.}) \\

			\hline
		\end{tabular}
	}
        \vspace{-0.2cm}
	\label{profile}
\end{table*}
\section{Settings for Evaluation}\label{sec:settings}
\subsection{Experimental Setting}
We conduct three scenarios to comprehensively evaluate the effectiveness of all approaches under different initial environmental resource densities: \textbf{Rich}, \textbf{Moderate}, and \textbf{Scarce}. The \textbf{Rich} environment indicates that 70\% of all resources required for player upgrades are spatially distributed throughout the game map and obtainable through tasks. Similarly, the \textbf{Moderate} and \textbf{Scarce} scenarios encompass 50\% and 30\% of the resources respectively. In MMOAgent setting, the parameter $m$ in Eq. \eqref{eq:write} for memory writing is 5 and the similarity threshold is 0.9 for importance accumulation. The $S$ in Eq. \eqref{eq:forget} is 20 and the importance score threshold for forgetting is 0.2.  For fair comparison, 10 agents with identical initial positions are placed in environments with uniformly distributed resources for each approach, simulated for 200 steps across 5 repetitions.
We use GPT-3.5-turbo provided by OpenAI API\footnote{\url{https://platform.openai.com/}} and Llama3-8B\footnote{\url{https://llama.meta.com/llama3/}}, a leading open-source LLM, as the LLM backbone of MMOAgent.

\subsection{Metrics}\label{sec:metrics}
We utilize two metrics as presented in Section \ref{exp} to evaluate the performance of the above agents. One is the agent's \textit{Capability} introduced in Section \ref{sec:env} which measures the agent's resource acquisition and management level and the higher the better. The other metric is \textit{Diversity}, which measures the variety of an agent's chosen activities by calculating activity distribution entropy: 
$H = -\sum_{i=1}^n p_i \log p_i$.
$n$ denotes the total count of activity categories and $p_i$ is the frequency with which the $i$-th activity category appears within the decision sequence. 
It is worth noting that neither an excessive nor a deficient level of diversity is optimal. Excessive diversity implies a lack of rules or advanced planning, while minimal diversity suggests being confined to specific activities without exploring alternative possibilities, indicating a limited understanding of the environment. The average \textit{Capability} and decision sequence \textit{Diversity} of 10 agents across 5 attempts are computed as the final performance.

\section{Profile Design}
\subsection{Realistic Player Features} 
\label{features}
We collect raw game logs from 16,294 players from March 4 to March 10, 2024. We implemented a series of essential privacy measures, including an industry-standard data collection protocol and a three-phase anonymization process to ensure original data irreversibility. After data preprocessing, we extracted multi-dimensional player features, including payment information, historical behavior, and game performance. Table \ref{feat_table} presents the main realistic player features utilized in Section \ref{profile_design} for constructing profiles.
\begin{table}
	\footnotesize
	\renewcommand\arraystretch{1.2}
	\centering
	\caption{Realistic player features name with descriptions. }
    \vspace{-0.3cm}
	{
		\begin{tabular}{p{2cm}|p{5.8cm}}
			\hline
			Feature Name& Descriptions \\ 
			\hline
			$online\_time$ & Player's average online hours per day in the game.\\
			$recharge\_money$ & The cumulative recharge amount of players in a week. \\
			$activity\_amount$ & Average daily in-game events participated by players. \\
			$rank\_in\_game$ & The player's ranking among all players.\\
			$role\_level$ & The level of the game character controlled by the player. \\
			$capacity$ &The number of game equipment owned by the player.\\
			$violations$ & Whether the player has been detected using cheats.  \\
           $vitality$ &The amount of energy consumed by the player in the game.  \\
			\hline
		\end{tabular}
	}	
	\label{feat_table}
\end{table}
\begin{table}
    \small
	\centering
	\caption{MMOAgen's performance under different memory parameters setting.}
    \vspace{-0.3cm}
	\renewcommand\arraystretch{1.0}
	{
		\begin{tabular}{l|c|c|c|c|c|c}
            \hline
			\multirow{2}[3]{*}{variants}   & \multicolumn{2}{c|}{Rich}  & \multicolumn{2}{c|}{Moderate}  & \multicolumn{2}{c}{Scarce}\\
   \cmidrule{2-7} 
			          & Cap. & Div.   & Cap. & Div.   & Cap. & Div.    \\
    \hline	
    MMOAgent & 121.0 &	1.5822 & 	80.4 	& 1.4149 	&75.0 & 	1.3106\\ 
   \hline
			STM size=2 & 117.2 &	1.5031 	&77.8 	&1.4052 &72.8 &1.2989    \\
			STM size=6 & 118.6 	&1.5427 &	79.2 &1.4147 &74.4 & 1.3096 	\\
			STM size=14  & 121.8 & 1.5874 &80.6 &1.4172 &74.2 &1.3081  \\
			STM size=18 & 122.2 &1.5944 &81.0 &1.4143 &73.6 &1.3078			 \\
			\hline
			
            LTM size=30 & 122.4 &1.5857 &80.2 &1.4146 &75.6 &1.3162\\ 
   \hline
			
	
			S=10 & 118.0 &1.5741 &77.8 &1.4109 &73.4 &1.3058 \\	
			S=30 & 117.2 &1.5688 &78.2 &1.4078 &74.0 &1.3117  \\	
			
            \hline
		\end{tabular}%
	}
	\vspace{-0.2cm}
	\label{memory_size}%
\end{table}
\subsection{Generated Profiles}\label{profile_text}
Utilizing GPT-4, we have synthesized summaries of five player profiles in Table \ref{profile}, focusing on analyzing the centroids of player characteristic clusters identified via k-means clustering.

\subsection{Evaluating the Human Consistency}\label{Evaluate}
\begin{figure}[H]
  \centering
  \includegraphics[width=0.91\linewidth]{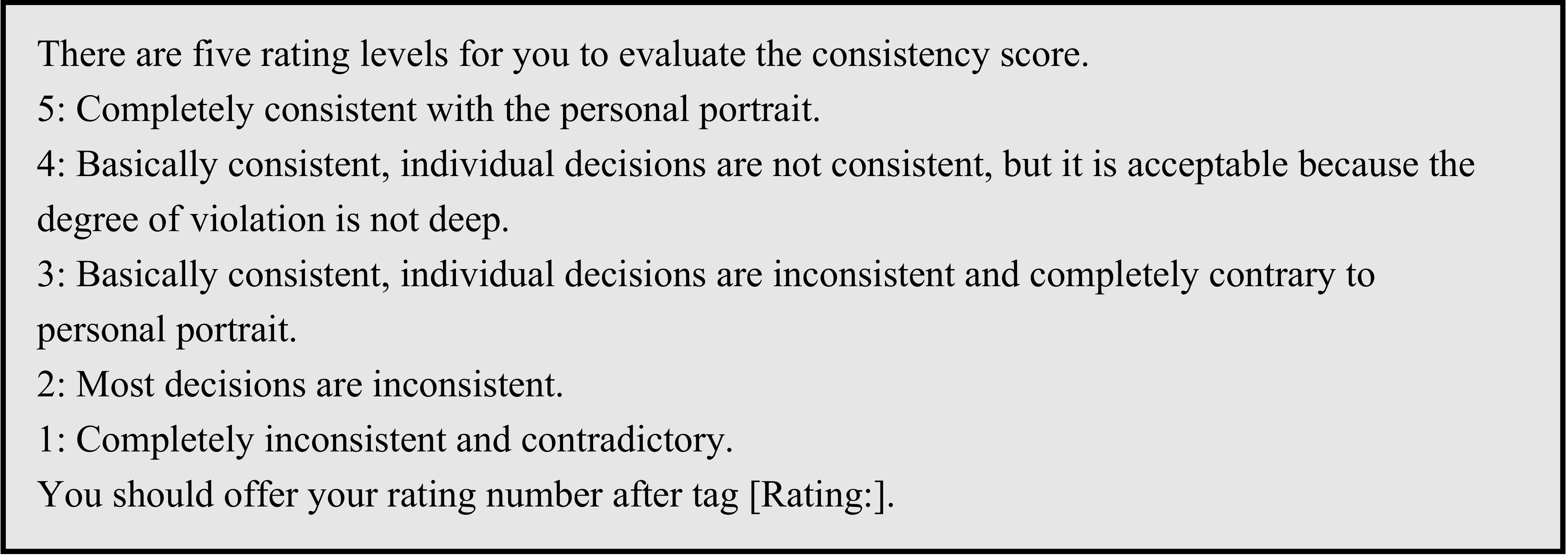}
\vspace{-0.3cm}
  \caption{The 5-tier rating system for evaluating the consistency between the agent's profile and its decision sequence.}
  \label{p-rating}
   \vspace{-0.4cm}
\end{figure}

For each agent, we collect a 1500-step-long decision sequence and cut it into subsequences with 15-timestep intervals. Then GPT-4 and human evaluators evaluate the subsequences based on the 5-tier rating system as shown in Figure \ref{p-rating}. The average evaluation scores of subsequences are calculated for each profile.

\section{Memory Influence}

We further investigate the influence of memory parameters introduced in Section \ref{memory} on MMOAgent's performance, focusing on the STM size, LTM size, and the memory decay control parameter S in eq. \eqref{eq:forget} , which are set as 10, 20, and 20 respectively in the paper. Results are listed in Table \ref{memory_size}, where "Cap." and "Div." denote Capability and Diversity. For STM size, which refer to the number of recent decisions in STM, we further test it with sizes of 2, 6, 14, 18. Results show an overall saturation trend. When STM size > 10, impact on agent performance is limited. In Scarce scenario, capability even drops slightly. This may be because the game environment changes in real time. The current state and a few recent decisions let the agent decide well. Too much historical info can mislead it. For LTM, due to limited memory capacity, memory forgetting helps the LTM retain more essential and general successful experiences, better emulating real human. Here, we further test S = 10, 30 and increased the LTM size from 20 to 30. Extreme S values, either too large or small, harm the model. A small S causes fast forgetting, losing crucial experiences. A large S means little forgetting, leaving outdated or unimportant experiences in LTM to mislead agent's decisions. Also, boosting LTM capacity barely improves MMOAgent's performance and even reduce it in Moderate scenario. This may because some unimportant experience from a bigger memory pool mislead the agent.

\balance
\end{document}